\documentclass[runningheads]{llncs}
\usepackage{graphicx}
\usepackage{amsmath,amssymb} % define this before the line numbering.
\usepackage{color}
\usepackage{graphicx}
\usepackage{amsmath,amssymb}
\usepackage{color}
\usepackage{tabularx}
\usepackage{multicol,multirow}
\usepackage{caption}
\usepackage{bm}
\usepackage{booktabs}
\usepackage{array}
\makeatletter
\def\hlinew#1{%
  \noalign{\ifnum0=`}\fi\hrule \@height #1 \futurelet
   \reserved@a\@xhline}
\makeatother

\begin{document}

%macro for raising the point in decimal numbers; see example in the abstract
\newcommand{\point}{
    \raise0.7ex\hbox{.}
    }

%Do   -- NOT --    use any additional macros

\pagestyle{headings}

\mainmatter

%===========================================================
\title{EdgeStereo: A Context Integrated Residual Pyramid Network for Stereo Matching} % Replace with your title

\titlerunning{EdgeStereo} % Replace with your title

\authorrunning{Xiao Song, Xu Zhao, Hanwen Hu, and Liangji Fang} % Replace with your names

\author{Xiao~Song, ~Xu~Zhao\thanks{Corresponding author. E-mail: song\_xiao@sjtu.edu.cn, zhaoxu@sjtu.edu.cn, huhanwen@sjtu.edu.cn, fangliangji@sjtu.edu.cn. This research is supported by the funding from NSFC (61673269,
61273285)}, ~Hanwen~Hu, and~Liangji~Fang} % Replace with your names

\institute{Department of Automation, Shanghai Jiao Tong University, China} % Replace with your institute's address

\maketitle

\begin{abstract}
Recent convolutional neural networks, especially end-to-end disparity estimation models, achieve remarkable performance on stereo matching task. However, existed methods, even with the complicated cascade structure, may fail in the regions of non-textures, boundaries and tiny details. Focus on these problems, we propose a multi-task network EdgeStereo that is composed of a backbone disparity network and an edge sub-network. Given a binocular image pair, our model enables end-to-end prediction of both disparity map and edge map. Basically, we design a context pyramid to encode multi-scale context information in disparity branch, followed by a compact residual pyramid for cascaded refinement. To further preserve subtle details, our EdgeStereo model integrates edge cues by feature embedding and edge-aware smoothness loss regularization. Comparative results demonstrates that stereo matching and edge detection can help each other in the unified model. Furthermore, our method achieves state-of-art performance on both KITTI Stereo and Scene Flow benchmarks, which proves the effectiveness of our design.
\end{abstract}

%===========================================================
\section{Introduction}

Stereo matching is a fundamental problem in computer vision. It has a wide
range of applications, such as robotics and autonomous driving \cite{schmid2013stereo,achtelik2009stereo}. Given a rectified image pair, the main goal is to find corresponding pixels from stereo images.
Most traditional stereo algorithms \cite{hirschmuller2005accurate,zhang2007estimating} follow the classical four-step pipeline \cite{scharstein2002taxonomy}, including matching cost computation, cost aggregation, disparity calculation and disparity refinement. However the hand-crafted features and multi-step regularized functions limit their improvements.

%such as \cite{hirschmuller2005accurate,zhang2007estimating}
Since \cite{zbontar2016stereo}, CNN based stereo methods extract deep features to represent image patches and compute matching cost. Although the performance on several benchmarks is significantly promoted, there remains some difficulties, including the limited receptive fields and complicated regularized functions.

\begin{figure}[tb]
\centering
\includegraphics[width=12.2cm]{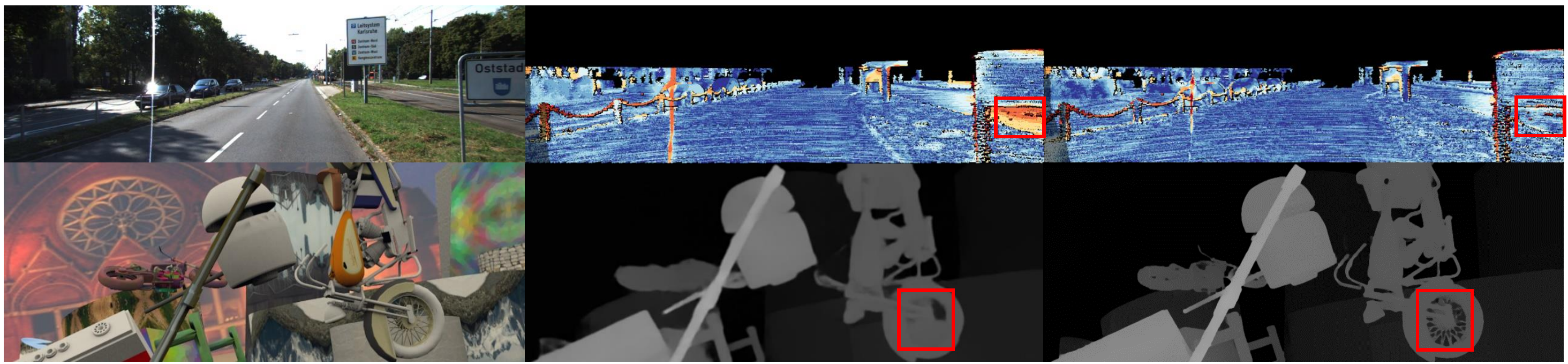}
\vspace{-3mm}
\caption{Examples on KITTI (top) and Scene Flow (bottom) datasets. Left: left stereo images. Middle up and right up: colorized error maps (wrong estimates in orange) predicted without context pyramid and by \emph{EdgeStereo}. Middle down and right down: colorized disparity map predicted without edge branch and by \emph{EdgeStereo}. As shown in the red boxes in top row, predicted disparities are more accurate in ill-posed regions such as shadowed road, under the guidance of context cues. In bottom row, \emph{EdgeStereo} produces accurate estimates in details under the guidance of edge cues.}
\label{intro}
\vspace{-2mm}
\end{figure}

%Driven by a large synthetic dataset \cite{mayer2016large},
Recently end-to-end disparity estimation networks \cite{mayer2016large,kendall2017end,pang2017cascade} achieve state-of-the-art performance, however drawbacks still exist. Firstly, it is difficult to handle local ambiguities in ill-posed regions. Secondly, the cascade structures or 3D convolution based structures are computationally expensive. Lastly, disparity predictions of thin structures or near boundaries are not accurate.

Humans, on the other hand, can find stereo correspondences easily by utilizing edge cues. Accurate edge contours can help discriminating between different objects or regions. In addition, humans perform binocular alignment well in texture-less or occluded regions based on global perception at different scales.

Based on these observations, we design a multi-task network \emph{EdgeStereo} that cooperates edge cues and edge regularization into disparity prediction pipeline. Firstly, we design a disparity network for \emph{EdgeStereo}, called context pyramid based residual pyramid network (\emph{CP-RPN}). Two modules are designed for \emph{CP-RPN}: a context pyramid to encode multi-scale context information for ill-posed regions, and an one-stage residual pyramid to simplify the cascaded refinement structure.
Secondly an edge detection sub-network is designed and employed in our unified model, to preserve subtle details with edge cues. Interactions between two tasks are threefold: (\romannumeral1) Edge features are embedded into disparity branch providing local and low-level representations. (\romannumeral2) The edge map, acting as an implicit regularization term, is fed to residual pyramid.  (\romannumeral3) The edge map is also utilized in \emph{edge-aware smoothness loss}, which further guides disparity learning.
%It breaks the routine that most disparity networks are simply modified from DispNetC or GC-Net without enough insights.

In disparity branch of \emph{EdgeStereo}, we use a siamese network with a correlation operation \cite{Dosovitskiy2015FlowNet} to extract image features and compute matching cost volumes, followed by a context pyramid. Based on different representations (unary features, edge features and matching cost volumes), context pyramid can encode contextual cues in multiple scales. Then they are aggregated as a hierarchical scene prior. Next we employ an hour-glass structure to regress the full-size disparity map. Different with the decoder in DispNetC or the cascade encoder-decoder in CRL, decoder in \emph{EdgeStereo} is replaced by the proposed residual pyramid. We predict the disparity map on the smallest scale and learn disparity residuals on other scales. Hence learning and refining are conducted in a single decoder, making \emph{CP-RPN} as an one-stage disparity estimation model. Based on experimental results in Table \ref{t1}, our residual pyramid is better and faster than other cascade structures. In edge branch of \emph{EdgeStereo}, the shallow part of backbone network is shared with \emph{CP-RPN}. Edge feature and edge map are embedded into the disparity branch, under the guidance of \emph{edge-aware smoothness loss}.

Both edge and disparity branches are fully-convolutional so that end-to-end training can be conducted for \emph{EdgeStereo}. As there is no dataset providing both ground-truth disparities and edge labels, we propose a multi-phase training strategy. We adopt the supervised disparity regression loss and our adapted edge-aware smoothness loss to train the entire \emph{EdgeStereo}, achieving a high accuracy on Scene Flow dataset \cite{mayer2016large}. We further finetune our model on KITTI 2012 and 2015 datasets, achieving state-of-the-art performance on KITTI stereo benchmarks. %We also make a large stereo dataset from KITTI raw data to further demonstrate the effectiveness of our design.
After multi-task learning, both disparity estimation and edge detection tasks are improved in both quantitatively and qualitatively.

In summary, our main contribution is threefold.

(\romannumeral1) We propose \emph{EdgeStereo} to support the joint learning of scene matching and edge detection, where edge cues and edge-aware smoothness loss serve as important guidance for disparity learning. The multi-task labels are not required during training due to our proposed multi-phase training strategy.

(\romannumeral2) The effective context pyramid is designed to handle ill-posed regions, and the efficient residual pyramid is designd to replace cascade refinement structures.

(\romannumeral3) Our unified model \emph{EdgeStereo} achieves state-of-the-art performance on Scene Flow dataset, KITTI stereo 2012 and 2015 benchmarks.

\section{Related Work}

\subsubsection{\emph{Stereo Matching.}} Among non-end-to-end deep stereo algorithms, each step in traditional stereo pipeline could be replaced by a network. For example, Luo \emph{et al.} \cite{luo2016efficient} train a simple multi-label classification network for matching cost computation. Shaked and Wolf \cite{shaked2016improved} introduce an initial disparity prediction network pooling global information from cost volume. Gidaris \emph{et al.} \cite{gidaris2016detect} substitute hand-crafted disparity refinement functions with a three-stage refinement network.

For end-to-end deep stereo algorithms, all steps in traditional stereo pipeline are combined for joint optimization. To train end-to-end stereo networks, Mayer \emph{et al.} \cite{mayer2016large} create a large synthetic stereo dataset, %making it possible to train a network for disparity estimation in an end-to-end manner.
meanwhile they also propose a baseline model called DispNet with an encoder-decoder structure. Based on DispNet, Pang \emph{et al.} \cite{pang2017cascade} cascade a residual learning network for further refinement.
%Liang \emph{et al.} \cite{liang2017learning} propose their iResNet including a different disparity refinement network.
Different from DispNet, Kendall \emph{et al.} \cite{kendall2017end} propose GC-Net that incorporates contextual information by means of 3D convolutions over a feature volume.
%in which the disparity map is regressed through a 3-D auto-encoder architecture ended with a ``soft-argmin" layer.
Based on GC-Net, Yu \emph{et al.} \cite{yu2018deep} add an explicit cost aggregation structure. An unsupervised method is proposed in \cite{zhong2017self}. Liang \emph{et al.} \cite{liang2017learning} formulate the disparity refinement task as Bayesian inference process for joint learning. PSMNet \cite{chang2018pyramid} utilizes spatial pyramid pooling and 3D CNN to regularize cost volumes. Our \emph{CP-RPN} is also an end-to-end network, but we explicitly encode context cues for disparity learning and our one-stage residual pyramid is efficient.

\vspace{-0.5cm}\subsubsection{\emph{Combining Stereo Matching with Other Tasks.}} Bleyer \emph{et al.} \cite{bleyer2011object} first solve stereo and object segmentation problems together. Guney and Geiger \cite{guney2015displets} propose Displets which utilizes foreground object recognition to help stereo matching. More tasks are fused through a slanted plane in \cite{yamaguchi2014efficient}. However, these hand-crafted multi-task methods are not robust.

\vspace{-0.5cm}\subsubsection{\emph{Edge Detection.}} To preserve details in disparity maps, we resort to edge detection task to supplement features and regularization. Inspired by FCN \cite{long2015fully}, Xie \emph{et al.} \cite{xie2015holistically} first design an  end-to-end edge detection network named holistically-nested edge detector (HED) based on VGG-$16$ network. %Next Liu \emph{et al.} \cite{liu2016learning} exploit relaxed labels.
Recently, Liu \emph{et al.} \cite{liu2017richer} modify the structure of HED, combining richer convolutional features from VGG backbone. These fully-convolutional edge networks can be easily incorporated with disparity estimation networks.

%Due to the usage of too many hand-crafted regularization functions, these methods are severely limited.

\vspace{-0.5cm}\subsubsection{\emph{Deep Learning Based Multi-task Structure.}} Cheng \emph{et al.} \cite{cheng2017segflow} propose an end-to-end network called SegFlow, which enables the joint learning of video object segmentation
and optical flow. The segmentation branch and flow branch are iteratively trained offline. Our \emph{EdgeStereo} is a different multi-task structure where multi-phase training is conducted rather than iterative training. Hence disparity branch can exploit more stable boundary information from pretrained edge branch. In addition, \emph{EdgeStereo} does not require multi-task labels from a single dataset, hence it is easier to find proper datasets for training.

\begin{figure}[tb]
\centering
\makebox[\textwidth][c]{\includegraphics[width=1.05\textwidth]{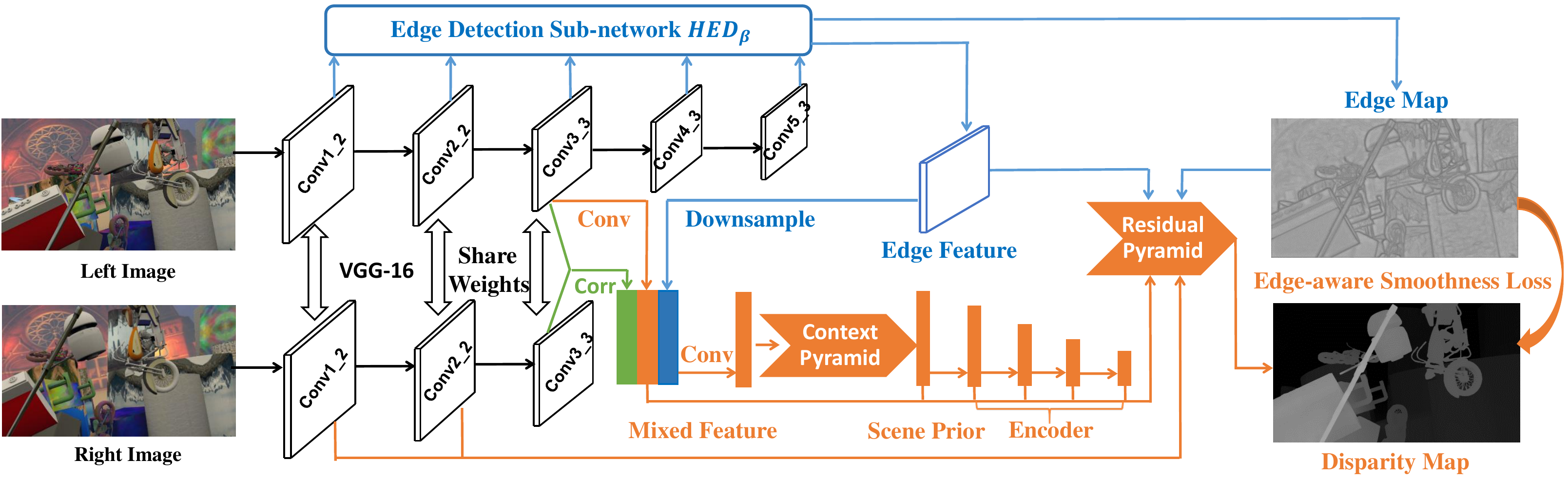}}
%\vspace{-5mm}
\caption{The overview of \emph{EdgeStereo}, consisting of a disparity network and an edge sub-network. They share the shallow part of backbone for effective computation. In disparity branch, context pyramid extract multi-scale context cues from the mixed feature representation. The hierarchical scene prior is encoded then decoded by our one-stage residual pyramid, producing a full-size disparity map. The edge branch cooperates edge cues into disparity estimation pipeline by embedding edge features and edge probability map. The edge map also guides disparity or residual learning under the guidance of \emph{edge-aware smoothness loss}. }
\label{archi}
\vspace{-2mm}
\end{figure}

\section{Approach}

In this section, we describe our multi-task model \emph{EdgeStereo}. We first present the basic
network structure. Then we introduce two critical modules in disparity branch: context pyramid and residual pyramid. Next we detail the cooperation strategies of edge cues, including edge feature embedding, edge map feeding and the adapted edge-aware smoothness loss. Finally we show how to conduct multi-task learning via our multi-phase training strategy.

\subsection{Basic Architecture}

The overall architecture of our \emph{EdgeStereo} is shown in Fig. \ref{archi}. To combine two tasks efficiently, edge branch shares the shallow computation with disparity branch at the backbone network. %Hence effective common features for two tasks can be learned.

%To compromise between fine-grained correspondence and large receptive field for matching cost calculation,
The shallow part of backbone network is used to extract image features $\mathbb{F}^l$ and $\mathbb{F}^r$ from the input pair, carrying local semantic information. Then a correlation layer in \cite{d2015flownet} is used to capture coarse correspondence between $\mathbb{F}^l$ and $\mathbb{F}^r$ in feature space, obtaining a cost volume $\mathbb{F}_c$. We also apply a convolution block on $\mathbb{F}^l$ to extract reduced image feature $\mathbb{F}_r^l$. Meanwhile in order to utilize representations with edge cues, we employ an edge sub-network to compute edge feature $\mathbb{F}_e^l$ from the reference image of disparity estimation (left image). The reduced image feature $\mathbb{F}_r^l$, the cost volume $\mathbb{F}_c$ and the edge feature $\mathbb{F}_e^l$ are concatenated then fused by an $1\times1$ convolution, forming the mixed feature representation $\mathbb{F}_m$.

Taking $\mathbb{F}_m$ as input, context pyramid collects contextual information at four scales and aggregate them into a hierarchical scene prior for disparity estimation. Each scale in context pyramid captures context cues from different sub-regions with different receptive fields. Next we feed the scene prior to an hour-glass structure to
predict full-size disparity map, where the encoder is a stack of convolution layers to sub-sample feature maps and the decoder is formulated by our residual pyramid. Multi-scale processing is conducted in residual pyramid, where the disparity map is directly regressed on the smallest scale and residual maps are predicted for refinement on other scales. Edge features and edge map are fed to each scale in residual pyramid, helping preserving details in disparity map. The edge map also guides disparity or residual learning under \emph{edge-aware smoothness loss regularization}. These are the key components of our framework. More settings are detailed in Section 4.1.
%by supplementing low-level representations and acting as an implicit regularization term.

\subsection{Context Pyramid}
Context information is widely used in many tasks \cite{mottaghi2014role,liu2015parsenet}. For stereo matching, it can be regarded as the relationship between an object and its surroundings or its sub-regions, which can help inferring correspondences especially for ill-posed regions. Many stereo methods learn these relationships by stacking lots of convolution blocks. Differently we encode context cues explicitly through context pyramid, hence learning stereo geometry of the scene is easier. Moreover, single-scale context information is insufficient because objects with arbitrary sizes are existed. Over-focusing on global information may neglect small-size objects, while disparities of big stuff might be inconsistent or discontinuous if the receptive field is small. Hence the proposed context pyramid aims at capturing multi-scale context cues in an efficient way.

We use four parallel branches with similar structures in the context pyramid. As mentioned in \cite{zhao2017pyramid}, the size of receptive field roughly indicates how much we use context. Hence four branches own different receptive fields to capture context information at different scales. The largest context scale corresponds to the biggest receptive field. To our knowledge, convolution, pooling and dilation \cite{chen2016deeplab} operations can enlarge the receptive field. Hence we design \emph{convolution context pyramid, pooling context pyramid and dilation context pyramid} respectively. They are detailed in Section 4.1. The best one is embedded in \emph{EdgeStereo}.

As shown in Fig. \ref{res}, outputs of four branches as well as the input $\mathbb{F}_m$ are concatenated as the hierarchical scene prior, carrying both low-level semantic information and global context cues for disparity estimation.

\subsection{Residual Pyramid}

Many stereo methods \cite{pang2017cascade,gidaris2016detect} use a cascade structure for disparity estimation, where the first network generates initial disparity predictions and the second network produces residual signals to rectify initial disparities. However these residual signals are hard to learn (residuals are always close to zero), because initial disparity predictions are pretty good. Moreover these multi-stage structures are computationally expensive.
In order to optimize the cascade structure, we design a residual pyramid so that initial disparity learning and disparity refining can be conducted in a single network.

\begin{figure}[tb]
\centering
\makebox[\textwidth][c]{\includegraphics[width=1.09\textwidth]{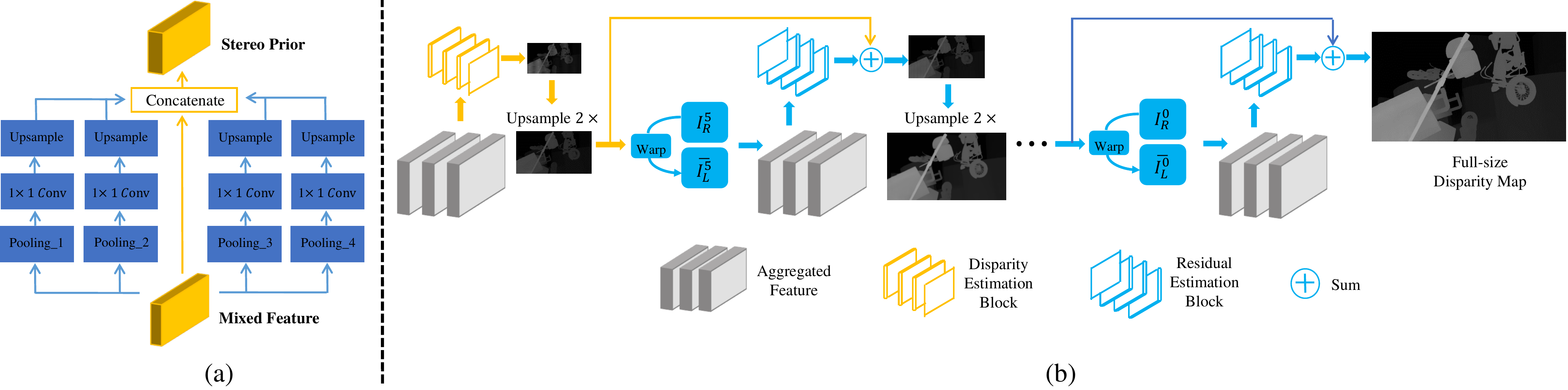}}
\vspace{-6mm}
\caption{(a) An example of pooling context pyramid. (b) One-stage residual pyramid. Disparity map is first predicted on the smallest scale then residual signals are predicted on other scales. Aggregated feature is the concatenation of edge cues, feature maps from encoder and geometrical constraints. Each estimation block is used to predict disparity or residual map, detailed in section 4.1.}
\label{res}
\vspace{-2mm}
\end{figure}

To make multi-scale disparity estimation easier, we refer to the idea ``From Easy to Tough'' from curriculum learning \cite{bengio2009curriculum}. In other words, it is easier to regress disparity map on the smallest scale because searching range is narrow and few details is needed. To get larger disparity maps, we estimate residual signals relative to the disparity map at the smallest scale. The formulation of residual pyramid makes \emph{EdgeStereo} an effective one-stage structure. Besides, the residual pyramid can be beneficial for overall training because it alleviates the problem of over-fitting.
%(\romannumeral3) At each scale besides the smallest, right image is downsampled then warpedaccording to the upsampled disparity map from previous scale, obtaining a synthesized left image. Error map between the synthesized and real left images is utilized as a geometric constraint for learning the residual map at each sacle.

Scale number $S$ in residual pyramid is consistent with the encoder structure. As shown in Fig. \ref{res}, the smallest scale in residual pyramid produces a disparity map $d_S$ ($\frac{1}{2^{S-1}}$ of the full resolution), then it is continuously upsampled and refined with the residual map $r_s$ on a larger scale, until the full-size disparity map $d_0$ is obtained. The formulation is shown in Eq. (\ref{sum}), where $u(\cdot)$ denotes upsampling by a factor of $2$ and $s$ denotes the pyramid scale (e.g. $0$ represents the full-resolution).

\vspace{-1.5mm}
\begin{equation}
d_s=u(d_{s+1})+r_s, \, 0\le s< S\,.
\label{sum}
\vspace{-0.5mm}
\end{equation}

For each scale, various information are aggregated to predict disparity or residual map, including the skip-connected feature map from encoder with higher frequency information, the edge feature and edge map (all interpolated to corresponding scale) to cooperate edge cues, and the geometrical constraints. For each scale except the smallest scale, we warp the resized right image $I_{R}^{s}$ according to disparity $d_s$ and obtain a synthesized left image $\overline{I_{L}^{s}}$. The error map $e_s=|I_{L}^{s}-\overline{I_{L}^{s}}|$ is a heuristic cue which can help to learn residuals. Hence the concatenation of $I_{L}^{s}$, $I_{R}^{s}$, $d_s$, $\overline{I_{L}^{s}}$ and $e_s$ serves as the geometrical constraints.

\subsection{Cooperation of Edge Cues}

Basic disparity estimation network \emph{CP-RPN} works well on ordinary and texture-less regions, where matching cues are clear or context cues can be easily captured through the context pyramid. However as shown in the second row of Fig. \ref{intro}, details in disparity map are lost, due to too many convolution and down-sampling operations. Hence we utilize edge cues to help refining disparity maps.

Firstly we cooperate edge cues by embedding edge features. On the one hand, in front of context pyramid, we combine the interpolated edge feature $\mathbb{F}_e^l$ with the image feature $\mathbb{F}_r^l$ and the cost volume $\mathbb{F}_c$. By concatenation, we expect context pyramid can consider both local semantic information, matching cost distribution and edge representations when extracting context cues. On the other hand, edge features are interpolated and concatenated to each scale in residual pyramid. This feature embedding alleviates the issue that residual pyramid lacks low-level representations to produce accurate disparities and residual signals.

Secondly we resize and feed the edge map to each scale in residual pyramid. The edge map acts as an implicit regularization term which can help smoothing disparities in non-edge regions and preserving edges in disparity map. Hence the edge sub-network does not behave like a black-box.

Finally we regularize the edge map into an edge-aware smoothness loss, which is an effective guidance for disparity estimation. For disparity smoothness loss $\mathcal{L}_{ds}$, we encourage disparities to be locally smooth and the loss term penalizes depth changes in non-edge regions. To allow for depth discontinuities at object contours, previous methods \cite{Godard2016Unsupervised,zhong2017self} weight this regularization term according to image gradients. Differently, we weight this term based on gradients of edge map, which is more semantically meaningful than intensity variation. As shown in Eq. (\ref{smoo}), $N$ denotes the number of pixels, $\partial d$ denotes disparity gradients and $\partial\mathcal{E}$ denotes gradients of edge probability map.

\vspace{-1.5mm}
\begin{equation}
\mathcal{L}_{ds}=\frac{1}{N}\sum_{i,j}|\partial_xd_{i,j}|e^{-|\partial_{x}\mathcal{E}_{i,j}|}+|\partial_yd_{i,j}|e^{-|\partial_{y}\mathcal{E}_{i,j}|}\,.
\label{smoo}
\vspace{-0.5mm}
\end{equation}

\vspace{-3mm}\subsection{Multi-phase Training Strategy and Objective Function}
In order to conduct multi-task learning for \emph{EdgeStereo}, we propose a multi-phase training strategy where the training phase is split into three phases. Weights of the backbone network are fixed in all three phases.

In the first phase, edge sub-network is trained on a dataset for edge detection task, guided by a class-balanced cross-entropy loss proposed in \cite{liu2017richer}.

In the second phase, we supervise the regressed disparities across $S$ scales on a stereo dataset. Deep supervision is adopted, forming the total loss as the sum $C=\sum_{s=0}^{S-1}C_s$ where $C_s$ denotes the loss at scale $s$. Besides the disparity smoothness loss, we adopt the disparity regression loss $\mathcal{L}_{r}$ for supervised learning, as shown in Eq. (\ref{re}).

\vspace{-1.5mm}
\begin{equation}
\mathcal{L}_{r}=\frac{1}{N}{||\,d-\hat{d}\,||}_1\,,
\label{re}
\vspace{-1.5mm}
\end{equation}
where $\hat{d}$ denotes the ground truth disparity map. Hence the overall loss at scale $s$ becomes $C_s=\mathcal{L}_{r}+\lambda_{ds}\mathcal{L}_{ds}$, where $\lambda_{ds}$ is a loss weight for smoothness loss. In addition, weights of the edge sub-network are fixed.

In the third phase, all layers in \emph{EdgeStereo} are optimized on the same stereo dataset used in the second phase. Similarly we also adopt the deep supervision across $S$ scales. However the edge-aware smoothness loss is not used in this phase, because edge contours in the second phase are more stable than those in the third phase. Hence the loss at scale $s$ is $C_s=\mathcal{L}_{r}$.

\section{Experiments}

%Experimental settings and results are presented in this section. To prove the effectiveness of our proposed CR-RPN and EdgeStereo, we conduct stereo matching ablation studies for them, visualize the better disparity and edge maps after multi-task learning, and compare with other state-of-the-art stereo methods.

Experiment settings and results are presented in this section. Firstly we evaluate key components of \emph{EdgeStereo} on Scene Flow \cite{mayer2016large} dataset. We also compare our approach with other state-of-the-art stereo matching methods on KITTI benchmarks. In addition, we demonstrate that better edge maps can be obtained after multi-task learning.

\subsection{Model Specifications}

The backbone network is VGG-16 \cite{Simonyan2014Very}. The shallow part of backbone shared by two tasks is conv1\_1 to conv3\_3. Hence the extracted unary features $\mathbb{F}^l$ and $\mathbb{F}^r$ have a $1/4$ spatial size to raw images. For cost volume computation, the max displacement in 1D-correlation layer is set to $40$.
%Then an $1\times1$ convolution block with $128$ channels is applied to the mixed feature representation $\mathbb{F}_m$ to help fusing various information before context pyramid.

The encoder contains convolution layers with occasional strides of $2$, resulting in a total down-sampling factor of $64$. Correspondingly there are $7$ output scales in residual pyramid. For each scale, the estimation block consists of four $3\times3$ convolution layers and the last convolution layer regresses disparity or residual map. Each convolution layer except the output is followed by a ReLU layer.

We modify the structure of HED \cite{xie2015holistically} and propose an edge sub-network called \emph{HED$_\beta$}, where low-level edge features are easier to obtain and the produced edge map is more semantic-meaningful. \emph{HED$_\beta$} uses the VGG-16 backbone from conv1\_1 to conv5\_3. In addition, we design five side branches from conv1\_2, conv2\_2, conv3\_3, conv4\_3 and conv5\_3 respectively. Each side branch consists of two $3\times3$ convolution layers, an upsampling layer and an $1\times1$ convolution layer producing the edge probability map. In the end, feature maps from each upsampling layer in each side branch are concatenated as the final edge feature, meanwhile edge probability maps in each side branch are fused as the final edge map. The final edge feature and edge map are of full size.

Finally we describe the structure of each context pyramid.

\emph{Convolution context pyramid.} Each branch consists of two convolution layers with a same kernel size. Kernel size for the largest context scale is biggest. For example, $7\times7, \, 5\times5, \, 3\times3$ and $1\times1$ for each branch, denoted as $C$-$7\_5\_3\_1$.

\emph{Pooling context pyramid.} Each branch consists of an average pooling layer with different pooling kernels, followed by an $1\times1$ convolution layer, then an upsampling layer to get the representation with a same spatial size as $\mathbb{F}_m$. For the largest context scale, output size of pooling layer is smallest. For example, $1\times1, \, 2\times2, \, 4\times4$ and $ 8\times8$ for each branch respectively, denoted as $P$-$1\_2\_4\_8$.

\emph{Dilation context pyramid.} Inspired by \cite{chen2016deeplab}, each branch consists of a $3\times3$ dilated convolution layer, followed by an $1\times1$ convolution layer to reduce dimensions. Dilation rate for the largest context scale is biggest. For example, $6,\,3,\,2$ and $1$ for each branch respectively, denoted as $D$-$6\_3\_2\_1$.

\subsection{Datasets and Evaluation Metrics}

Scene Flow dataset \cite{mayer2016large} is a synthesised dataset containing $35454$ training and $4370$ test image pairs. Dense ground-truth disparities are provided and we perform the same screening operation as CRL \cite{pang2017cascade}. The real-world KITTI dataset includes two subsets with sparse ground-truth disparities. KITTI 2012 \cite{geiger2012we} contains $194$ training and $195$ test image pairs while KITTI 2015 \cite{menze2015object} consists of $200$ training and $200$ test image pairs.
%We split two KITTI validation sets with $40$ image pairs each from two KITTI training sets.

To pretrain the edge sub-network, we adopt the BSDS500 \cite{arbelaez2011contour} dataset containing $300$ training and $200$ test images. Consistent with \cite{liu2016learning,liu2017richer}, we combine the training data in BSDS500 with PASCAL VOC Context dataset \cite{mottaghi2014role}.

To evaluate the stereo matching results, we apply the end-point-error (EPE) which measures the average Euclidean distance between estimated and ground-truth disparity. We also use the percentage of bad pixels whose disparity errors  are greater than a threshold ($>t\,px$), denoted as $t$-pixel error.

\begin{table}[tb]
\scriptsize
\renewcommand{\captionfont}{\scriptsize}
\centering
\caption{Ablation studies on Scene Flow dataset. $\mathbb{F}_h$ denotes the hybrid feature, which is the aggregation of unary feature and cost volume then fused by an $1\times1$ convolution. $\mathbb{F}_h$ is equivalent to the mixed feature $\mathbb{F}_m$ without edge feature embedding. $\mathbb{F}_h$ is only used in pure disparity networks.}
\label{t1}
%\vspace{-0.2cm}
\begin{tabularx}{12.2cm}{p{8.28cm}|p{1.4cm}<{\centering}|X<{\centering}|X<{\centering}}
\hlinew{0.75pt}
\specialrule{0em}{0.5pt}{0.5pt}
%\multirow{2}{*}{Model} & \multirow{2}{*}{$>3\,px (\%)$} & \multirow{2}{*}{EPE} & \multirow{2}{*}{Time(s)}  \\
%& & &\\
Model & $>3\,px (\%)$ & EPE & Time(s)  \\
\specialrule{0em}{0.5pt}{0.5pt}
\hline
\specialrule{0em}{0.5pt}{0.5pt}
\multicolumn{4}{c}{Basic Ablation Studies}\\
\specialrule{0em}{0.5pt}{0.5pt}
\hline
%DispNetC \cite{mayer2016large} &  & 9.67 &  & 1.84 & 60 \\
DispFulNet \cite{pang2017cascade} (unary feature + encoder-decoder) & 8.61 & 1.75 & 0.07 \\
Our unary feature + encoder-decoder in DispFulNet  & 6.83 & 1.38 & 0.14 \\
$\mathbb{F}_h$ + encoder-decoder in DispFulNet & 6.70  & 1.37 & 0.14 \\
$\mathbb{F}_h$ + our encoder-decoder (residual pyramid)  & 5.96 & 1.28 & 0.19 \\
$\mathbb{F}_h$ + $P$-$2\_4\_8\_16$ + our encoder-decoder (\emph{CP-RPN}) & \textbf{5.33} & \textbf{1.15} & 0.19 \\
$\mathbb{F}_h$ + $P$-$2\_4\_8\_16$ + encoder-decoder in DispFulNet & 5.95  & 1.28 & 0.13 \\
\hline
\specialrule{0em}{0.5pt}{0.5pt}
\multicolumn{4}{c}{Context Pyramid Comparisons}\\
\specialrule{0em}{0.5pt}{0.5pt}
\hline
\emph{CP-RPN} with convolution pyramid $C$-$7\_5\_3\_1$ & 5.85& 1.26 & 0.24 \\
\emph{CP-RPN} with convolution pyramid $C$-$9\_7\_5\_3$ & 5.70 & 1.21 & 0.26 \\
\emph{CP-RPN} with convolution pyramid $C$-$11\_9\_7\_5$ & 5.79 & 1.23 & 0.28 \\
\emph{CP-RPN} with pooling pyramid $P$-$1\_2\_4\_8$ & 5.61 & 1.19 & 0.22 \\
\emph{CP-RPN} with pooling pyramid $P$-$2\_4\_8\_16$ & \textbf{5.33} & \textbf{1.15} & \textbf{0.19} \\
\emph{CP-RPN} with dilation pyramid $D$-$6\_3\_2\_1$ & 5.81 & 1.24 & 0.23 \\
\emph{CP-RPN} with dilation pyramid $D$-$12\_9\_6\_3$ & 5.52 & 1.17 & 0.24 \\
\emph{CP-RPN} with dilation pyramid $D$-$24\_18\_12\_6$ & 5.88  & 1.26 & 0.24 \\
\hline
\specialrule{0em}{0.5pt}{0.5pt}
\multicolumn{4}{c}{One-stage \emph{vs} Multi-stage Refinement}\\
\specialrule{0em}{0.5pt}{0.5pt}
\hline
$\mathbb{F}_h$ + $P$-$2\_4\_8\_16$ + encoder-decoder in DispFulNet + DRR \cite{gidaris2016detect} & 5.48 & 1.17 & 0.47\\
$\mathbb{F}_h$ + $P$-$2\_4\_8\_16$ + encoder-decoder and refinement in CRL \cite{pang2017cascade} & 5.34 & 1.16 & 0.31 \\
$\mathbb{F}_h$ + $P$-$2\_4\_8\_16$ + our one-stage encoder-decoder & \textbf{5.33} & \textbf{1.15} & \textbf{0.19} \\
\hline
\specialrule{0em}{0.5pt}{0.5pt}
\multicolumn{4}{c}{Benefits from Edge Cues}\\
\specialrule{0em}{0.5pt}{0.5pt}
\hline
\emph{CP-RPN} with $C$-$7\_5\_3\_1$ & 5.85 & 1.26 & 0.24 \\
\emph{CP-RPN} with $C$-$7\_5\_3\_1$ + edge cues & 5.40 & 1.14 & 0.32 \\
$\mathbb{F}_h$ + $D$-$6\_3\_2\_1$ + encoder-decoder in DispFulNet & 6.31 & 1.33 & 0.17 \\
$\mathbb{F}_m$ + $D$-$6\_3\_2\_1$ + encoder-decoder in DispFulNet + edge cues & 5.98 & 1.27 & 0.25 \\
$\mathbb{F}_h$ + $P$-$2\_4\_8\_16$ + our encoder-decoder & 5.33 & 1.15 & 0.19 \\
$\mathbb{F}_m$ + $P$-$2\_4\_8\_16$ + our encoder-decoder + edge cues (\emph{EdgeStereo}) & \textbf{4.97} & \textbf{1.11} & 0.29 \\
\hlinew{0.75pt}
\end{tabularx}
\vspace{-0.5cm}
\end{table}

\subsection{Implementation Details} Our model is implemented based on Caffe \cite{jia2014caffe}. The model is optimized using the Adam method \cite{kingma2014adam} with $\beta_1=0.9$, $\beta_2=0.999$.
In the first training phase, \emph{HED$_\beta$} is trained on BSDS500 dataset for $30k$ iterations. The batch size is $12$ and the initial learning rate is $10^{-6}$ which is divided by $10$ at the $15k$-th and $25k$-th iterations. The second and third training phases are all conducted on Scene Flow dataset with a batch size of $2$. In the second phase, we train for $400k$ iterations with a fixed learning rate of $10^{-4}$. The loss weight $\lambda_{ds}$ for edge-aware smoothness loss is set to $0.1$.  Afterwards in the third phase, we train for $600k$ iterations with a learning rate of $10^{-4}$ which is halved at the $300k$-th and $500k$-th iterations.
When finetuning on KITTI datasets, the initial learning rate is set to $2\times10^{-5}$ which is halved at the $20k$-th and $80k$-th iterations. Since ground-truth disparities provided by the KITTI datasets are sparse, invalid pixels are neglected in $\mathcal{L}_{r}$.

\subsection{Ablation Studies}
In this section, we conduct several ablation studies on Scene Flow dataset to evaluate key components in the \emph{EdgeStereo} model.
%including local stereo volume extraction, various context pyramid and residual pyramid in CP-RPN, and the role of HED$_\beta$ in EdgeStereo compared to the pure CP-RPN.
The one-stage DispFulNet \cite{pang2017cascade} (a simple variant of DispNetC \cite{mayer2016large}) serves as the baseline model in our experiments. %DispFulNet consists of unary feature extraction, 1-D correlation and the encoder-decoder.
All results are shown in Table \ref{t1}.

\vspace{-0.5cm}\subsubsection{\emph{Hybrid Feature Extraction.}}
%local stereo volume is used to provide local semantic representations and coarse correspondence distributions for further disparity regression.
Firstly we replace the unary feature extraction part in DispFulNet with the shallow part of VGG-16 bcakbone, $3$-pixel error is reduced from $8.61\%$ to $6.83\%$. Next we apply an $1\times1$ convolution with $128$ channels on the concatenation of unary feature and cost volume, forming the hybrid feature $\mathbb{F}_h$. The $3$-pixel error is further reduced to $6.70\%$ because it can help fusing various information such as local semantic features and matching cost distributions. For clarification, hybrid feature $\mathbb{F}_h$ is equivalent to the mixed feature $\mathbb{F}_m$ in Section 3.1 without edge feature embedding.

\vspace{-0.5cm}\subsubsection{\emph{Context Pyramid.}} Firstly we choose a context pyramid ($P$-$2\_4\_8\_16$), then train a model consisting of the hybrid feature extraction part, the selected context pyramid and the encoder-decoder of DispFulNet. Compared with the model without context pyramid, $3$-pixel error is reduced from $6.70\%$ to $5.95\%$. Furthermore, as shown in the ``Context Pyramid Comparisons'' part in Table \ref{t1}, adopting other context pyramids can also lower the $3$-pixel error. Hence we argue that multi-scale context cues are beneficial for dense disparity estimation task.

\vspace{-0.5cm}\subsubsection{\emph{Encoder-Decoder (Residual Pyramid).}} We use the same encoder as DispFulNet and we adopt residual pyramid as the decoder. To prove its effectiveness, we train a model consisting of the hybrid feature extraction part and our encoder-decoder. Compared with the model containing the encoder-decoder in DispFulNet, $3$-pixel error is reduced from $6.70\%$ to $5.96\%$.
Hence our multi-scale residual learning mechanism is superior to direct disparity regression.

Finally we train \emph{CP-RPN} consisting of the hybrid feature extraction part, context pyramid $P$-$2\_4\_8\_16$ and our encoder-decoder. The $3$-pixel error is $5.33\%$ and the EPE is $1.15$, outperforming the baseline model by $3.28\%/0.60$.
%Hence all major components in CP-RPN can help the disparity estimation.

\vspace{-0.5cm}\subsubsection{\emph{Context Pyramid Comparisons.}} %In Sec.3, three kinds of context pyramid are proposed.
We train different \emph{CP-RPN} models with different context pyramids. As shown in Table \ref{t1}, convolution context pyramids don't work well, reducing the $3$-pixel error by only $0.11\%$, $0.26\%$ and $0.17\%$ respectively. In addition, the large dilation rate is harmful for extracting context cues. The $3$-pixel error of $D$-$12\_9\_6\_3$ is $5.52\%$ while $5.88\%$ for $D$-$24\_18\_12\_6$. $P$-$2\_4\_8\_16$ has the best performance, achieving a $3$-pixel error of $5.33\%$. Hence pooling context pyramid $P$-$2\_4\_8\_16$ is embedded in the final model.

\vspace{-0.5cm}\subsubsection{\emph{Comparisons with Multi-stage Refinement.}} Firstly we compare with the three-stage refinement structure DRR \cite{gidaris2016detect}. We replace our encoder-decoder with encoder-decoder in DispFulNet, then three additional networks are cascaded for refinement. \emph{CP-RPN} outperforms this model by $0.15\%$ meanwhile being $2.3$ times faster. Next we compare with the two-stage cascade structure CRL \cite{pang2017cascade}. We replace our encoder-decoder with disparity prediction and disparity refinement networks in \cite{pang2017cascade}.  As can be seen, performance is almost equal but our model is faster with less parameters, which proves the effectiveness of residual pyramid.

\vspace{-0.5cm}\subsubsection{\emph{Benefits from Edge Cues.}} We conduct several experiments where different disparity networks are cooperated with our edge sub-network. As can be seen, all stereo matching models are improved. We also present visual demonstrations as shown in Fig. \ref{disp1}. When edge cues are cooperated into the disparity estimation pipeline, subtle details are preserved hence the error rate is reduced.

\begin{table}[tb]
\scriptsize
%\vspace*{-7pt}
\renewcommand{\captionfont}{\scriptsize}
\centering
\caption{Comparisons of stereo matching methods on Scene Flow dataset.}
%\vspace{-0.2cm}
\label{t2}
\begin{tabularx}{12.2cm}{p{1cm}<{\centering}||X<{\centering}|X<{\centering}|X<{\centering}|X<{\centering}|X<{\centering}|X<{\centering}|X<{\centering}|X<{\centering}|X<{\centering}|X<{\centering}}
\hlinew{0.75pt}
\multirow{2}{*}{metric} & \multirow{2}{*}{SGM} & \multirow{2}{*}{SPS-st} & MC- & \multirow{2}{*}{DRR} & \multirow{2}{*}{DispNet} & DispFul & \multirow{2}{*}{CRL} & \multirow{2}{*}{GC-Net} & \multirow{2}{*}{CA-Net} &  Edge \\
& &  & CNN & &  & Net & & & & Stereo \\
\hline
$>3\,px$ & 12.54 & 12.84 & 13.70 & 7.21 & 9.67 & 8.61 & 6.20 & 7.20 & 5.62 & \textbf{4.97} \\
\hline
EPE & 4.50 & 3.98 & 3.79 & - & 1.84 & 1.75 & 1.32 & - & - & \textbf{1.11} \\
\hlinew{0.75pt}
\end{tabularx}
%\vspace{-0.4cm}
\end{table}

\subsection{Comparisons with Other Stereo Methods}
In this section, we compare \emph{EdgeStereo} with state-of-the-art stereo matching methods on Scene Flow dataset as well as KITTI 2012 and 2015 benchmarks.
\vspace{-0.5cm}\subsubsection{\emph{Scene Flow Results.}} 
\begin{figure}[tb]
\centering
\includegraphics[width=12.2cm]{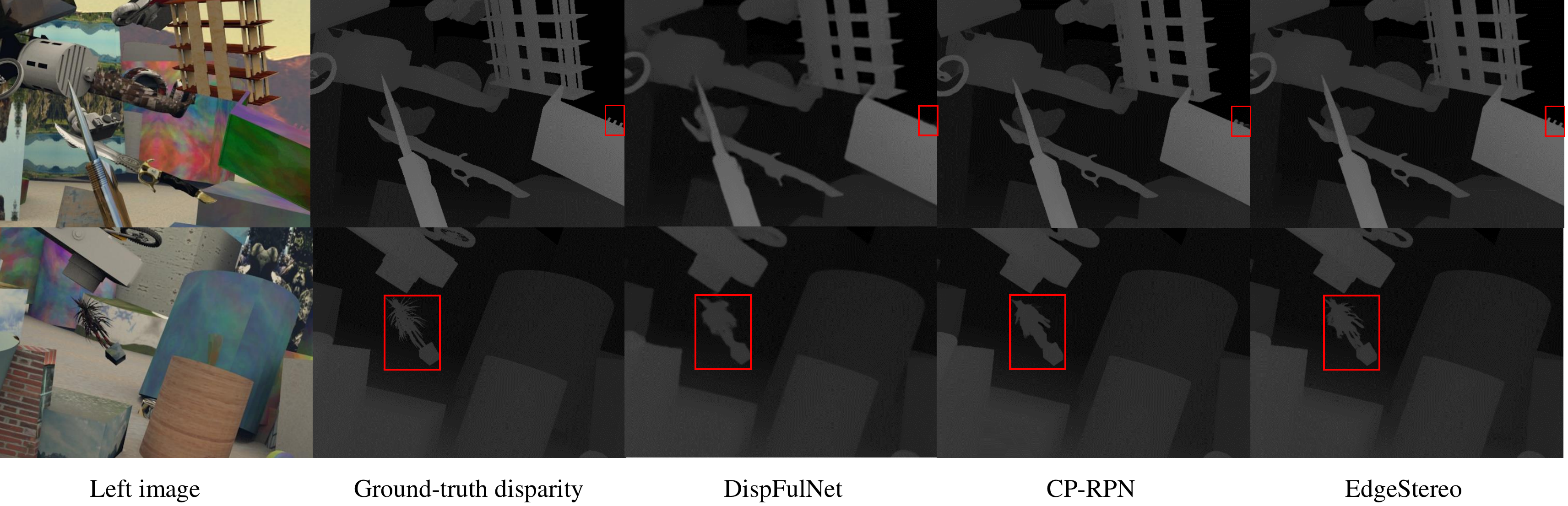}
%\vspace{-4mm}
\caption{Comparisons of different stereo models on Scene Flow dataset.}
\label{disp1}
\vspace{-2mm}
\end{figure}

Firstly we compare with several non-end-to-end methods, including SGM \cite{hirschmuller2005accurate}, SPS-St \cite{yamaguchi2014efficient}, MC-CNN-fst \cite{zbontar2016stereo} and DRR \cite{gidaris2016detect}. We also compare with the most advanced end-to-end stereo networks, including DispNetC \cite{mayer2016large}, DispFulNet \cite{pang2017cascade}, CRL \cite{pang2017cascade}, GC-Net \cite{kendall2017end} and CA-Net \cite{yu2018deep}. The comparisons are presented in Table \ref{t2}, \emph{EdgeStereo} achieves the best performance in terms of two evaluation metrics. As shown in Fig. \ref{disp1}, disparities predicted by \emph{EdgeStereo} are very accurate, especially in thin structures and near boundaries.

\vspace{-0.5cm}\subsubsection{\emph{KITTI Results.}}

For KITTI 2012,  \emph{EdgeStereo} is finetuned on all $194$ training image pairs, then test results are submitted to KITTI stereo 2012 benchmark. For evaluation, we use the percentage of erroneous pixels in non-occluded (Noc) and all (All) regions. We also conduct comparisons in challenging reflective (Refl) regions such as car windows. The results are shown in Table \ref{t3}. By leveraging context and edge cues, our \emph{EdgeStereo} model is able to handle challenging scenarios with large occlusion, texture-less regions and thin structures.

\begin{figure}[tb]
\centering
\includegraphics[width=12.2cm]{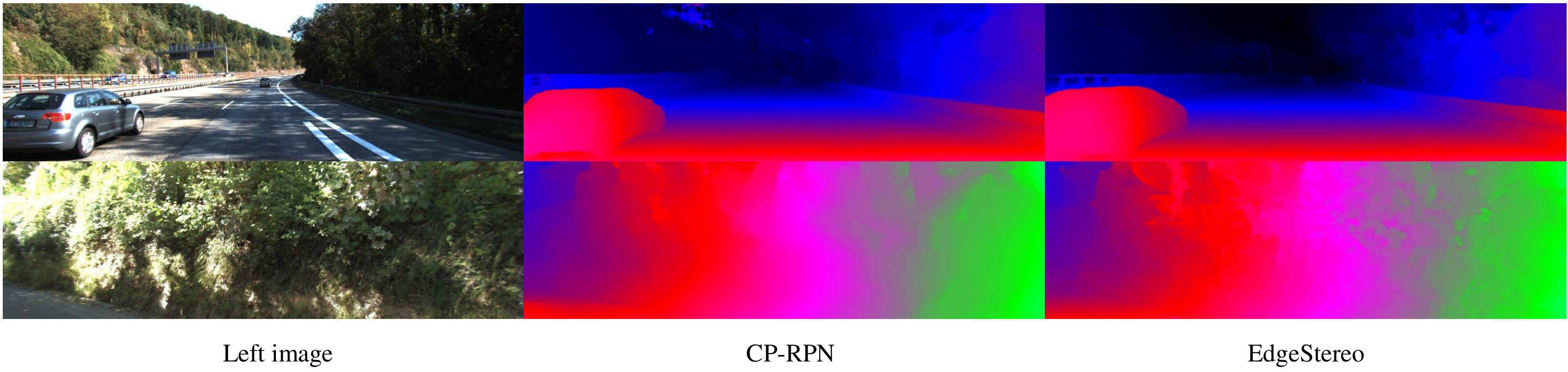}
%\vspace{-4mm}
\caption{Benefits from edge branch on KITTI dataset. As can be seen, after incorporating edge cues, predicted disparities are more accurate in thin structures, near boundaries and the upper part of the images.}
\label{kitti_edge}
\vspace{-1mm}
\end{figure}

\begin{figure}[tb]
%\vspace{-5mm}
\centering
\includegraphics[width=\textwidth]{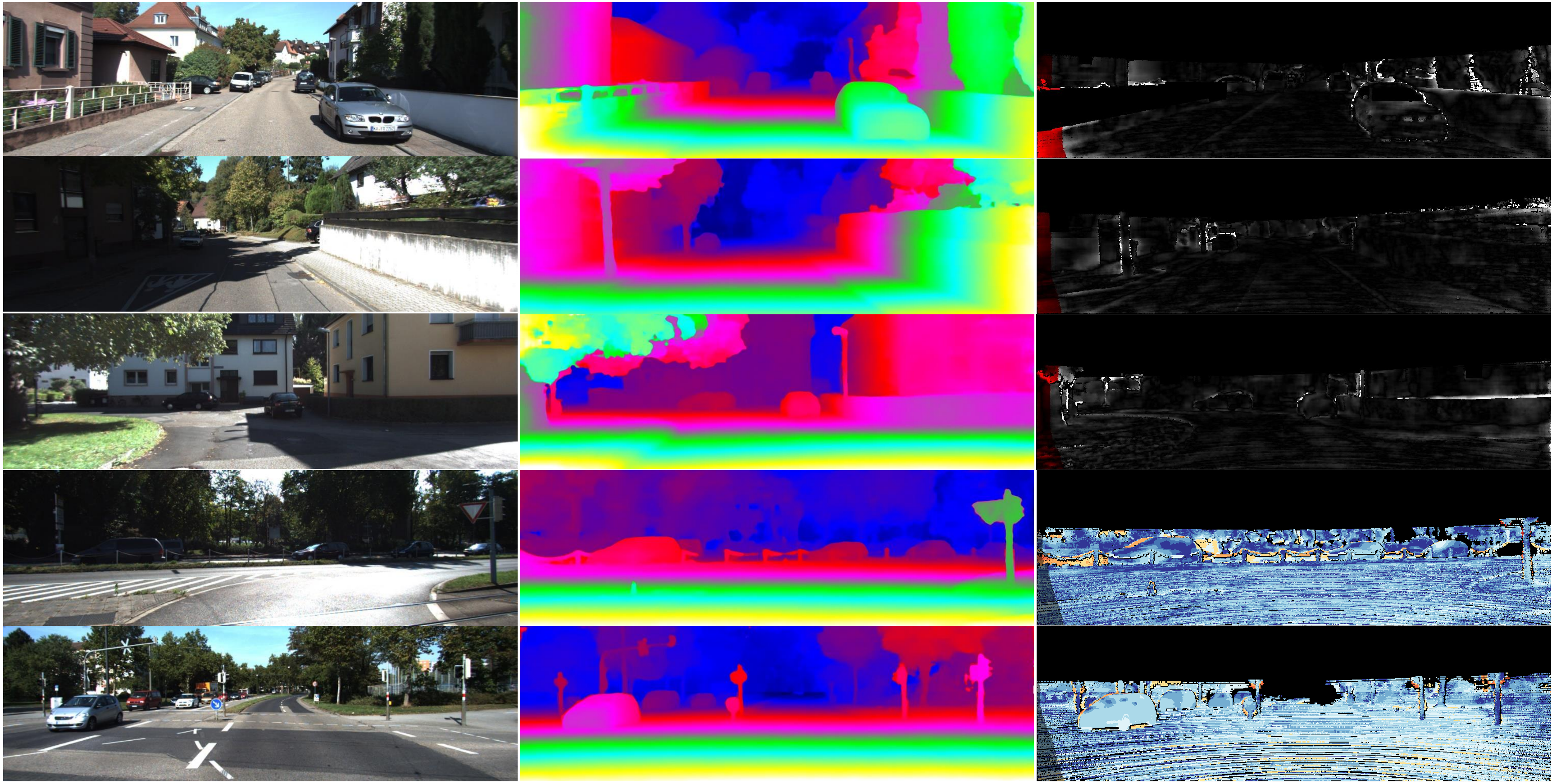}
\caption{Qualitative results on the KITTI datasets. The top three rows are from KITTI 2012 test set and the following rows are from KITTI 2015 test set. From left: left stereo input image, disparity prediction, error map.}
\label{kitti}
\end{figure}

For KITTI 2015, we also finetune \emph{EdgeStereo} on the whole training set. The test results are also submitted. For evaluation, we use the $3$-pixel error of background (D1-bg), foreground (D1-fg) and all pixels (D1-all) in non-occluded and all regions. The results are shown in Table \ref{t4}. \emph{EdgeStereo} achieves state-of-the-art performance on KITTI 2015 benchmark and our one-stage structure is faster than most stereo models. Fig. \ref{kitti} gives qualitative results on KITTI test sets. As can be seen, \emph{EdgeStereo} produces high-quality disparity maps in terms of global scene and object details. We also provide visual demonstrations of ``stereo benefits from edge'' on KITTI datasets, as shown in Fig. \ref{kitti_edge}.

\begin{figure}[tb]
\vspace{-1mm}
\centering
\includegraphics[width=10.2cm]{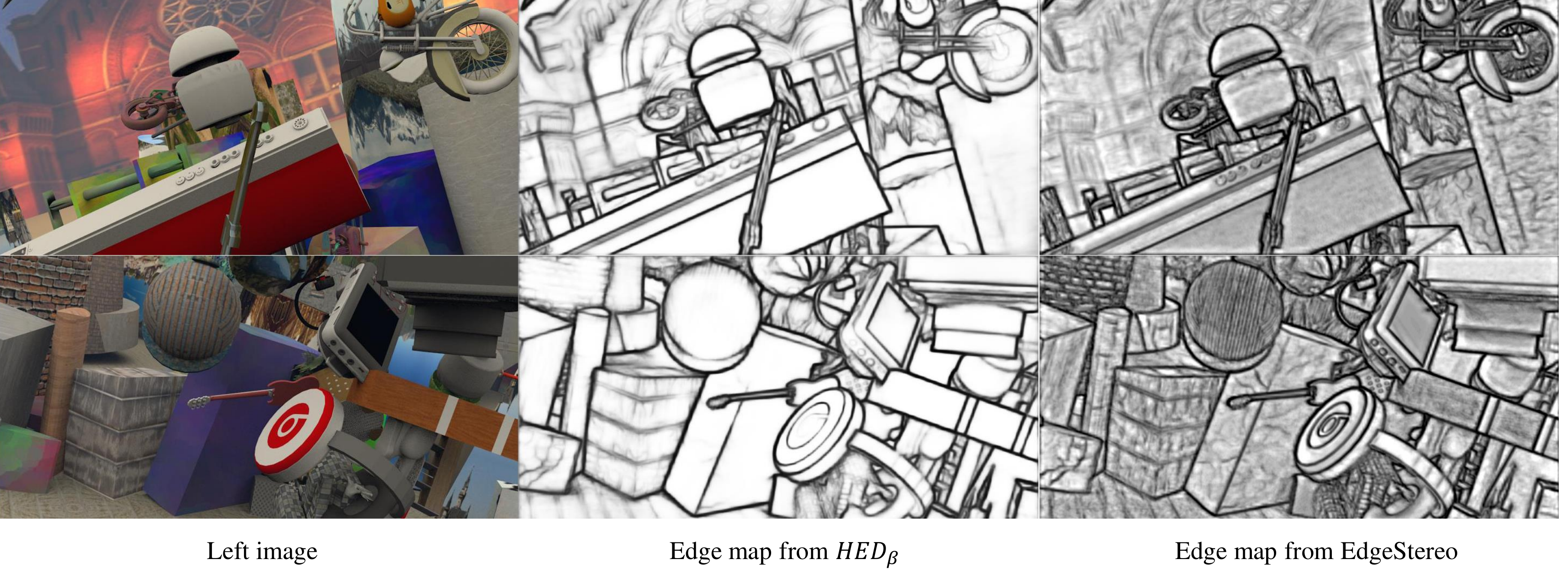}
%\vspace{-1mm}
\caption{Visual demonstrations for better edge maps after multi-task learning. As can be seen, details are highlighted in the produced edge maps.}
\label{edge}
\vspace{-4mm}
\end{figure}

\subsection{Better Edge Map}
We can't evaluate on a stereo dataset whether the edge detection task is improved or not after multi-task learning, because the ground-truth edge map is not provided. Hence we first give visual demonstrations on Scene Flow dataset, as shown in Fig. \ref{edge}. \emph{EdgeStereo} produces edge maps with finer details, compared with \emph{HED$_\beta$} without multi-task learning. We argue that the learned geometrical knowledge from disparity branch can help highlighting image boundaries.

For quantitative demonstrations, we conduct further experiments on BSDS500 dataset. ODS F-measure \cite{liu2017richer} (higher is better) is $0.790$ for original \emph{HED$_\beta$}, $0.795$ for \emph{HED$_\beta$} after multi-task learning and $0.788$ for the baseline model HED. All models are finetuned on BSDS500 dataset for same epochs.

\section{Conclusion}
In this paper, we present a multi-task architecture \emph{EdgeStereo} where edge cues are incorporated into the disparity estimation pipeline. Also the proposed context pyramid and residual pyramid enable our unified model to handle challenging scenarios with an effective one-stage structure. Our method achieves state-of-the-art performance on Scene Flow dataset and KITTI stereo benchmarks, demonstrating the effectiveness of our design.

%Experimental results show that proposed EdgeStereo achieves state-of-the-art performance on two public stereo benchmarks. In addition, both stereo matching task and edge detection task get better after the multi-task learning.

\begin{table}[tb]
\scriptsize
\renewcommand{\captionfont}{\scriptsize}
\centering
\caption{Results on the KITTI stereo 2012 benchmark. The online leaderboard ranks all methods according to the $3$-pixel error in ``Noc'' regions.}
\label{t3}
%\vspace{-0.25cm}
\begin{tabularx}{12.2cm}{p{2.45cm}|X<{\centering}X<{\centering}X<{\centering}|X<{\centering}X<{\centering}X<{\centering}|X<{\centering}X<{\centering}X<{\centering}}
\hlinew{0.75pt}
& \multicolumn{3}{c|}{$>\,3$px} & \multicolumn{3}{c|}{$>\,4$px} & \multicolumn{3}{c}{$>\,5$px} \\
%\cline{2-9}
& \textbf{\emph{Noc}} & All & Refl & Noc & All & Refl & Noc & All & Refl \\
\hline
PSMNet \cite{chang2018pyramid} & \textbf{1.49} & \textbf{1.89} & 8.36 & \textbf{1.12} & \textbf{1.42} & 5.89 & \textbf{0.90} & \textbf{1.15} & 4.58 \\
iResNet \cite{liang2017learning} & 1.71 & 2.16 & 7.40 & 1.30 & 1.63 & 5.07 & 1.06 & 1.32 & 3.82 \\
GC-Net \cite{kendall2017end} & 1.77 & 2.30 & 10.80 & 1.36 & 1.77 & 8.13 & 1.12 & 1.46 & 6.59 \\
L-ResMatch \cite{shaked2016improved} & 2.27 & 3.40 & 15.94 & 1.76 & 2.67 & 12.92 & 1.50 & 2.26 & 11.14\\
SGM-Net \cite{seki2017sgm} & 2.29 & 3.50 & 15.31 & 1.83 & 2.80 & 12.18 & 1.60 & 2.36 & 10.39 \\
SsSMNet \cite{zhong2017self} & 2.30 & 3.00 & 14.02 & 1.82 & 2.39 & 10.87 &1.53 &2.01 & 8.96 \\
PBCP \cite{seki2016patch} & 2.36 & 3.45 & 16.78 & 1.88 & 2.28 & 13.40 & 1.62 & 2.32 & 11.38\\
Displets v2 \cite{guney2015displets} & 2.37 & 3.09 & 8.99 & 1.97 & 2.52 & 6.92 & 1.72 & 2.17 & 5.71 \\
MC-CNN-acrt \cite{zbontar2016stereo} & 2.43 & 3.63 & 17.09 & 1.90 & 2.85 & 13.76 & 1.64 & 2.39 & 11.72 \\
\hline
EdgeStereo (ours) & 1.73 & 2.18 & \textbf{7.01} & 1.30 & 1.64  & \textbf{4.83} & 1.04 & 1.32 & \textbf{3.73}\\
\hlinew{0.75pt}
\end{tabularx}
\end{table}

\begin{table}[tb]
\scriptsize
\renewcommand{\captionfont}{\scriptsize}
\centering
\caption{Results on the KITTI stereo 2015 benchmark. The online leaderboard ranks all methods according to the D1-all error of ``All Pixels''.}
\label{t4}
%\vspace{-0.25cm}
\begin{tabularx}{12.2cm}{p{2.45cm}|X<{\centering}X<{\centering}X<{\centering}|X<{\centering}X<{\centering}X<{\centering}|X<{\centering}}
\hlinew{0.75pt}
& \multicolumn{3}{c|}{All Pixels} & \multicolumn{3}{c|}{Non-Occluded Pixels} & Runtime \\
%\cline{2-9}
& D1-bg & D1-fg & \textbf{\emph{D1-all}} & D1-bg & D1-fg & D1-all & (s) \\
\hline
PSMNet \cite{chang2018pyramid} & \textbf{1.86} & 4.62 & \textbf{2.32} & \textbf{1.71} & 4.31 & \textbf{2.14} & 0.41\\
iResNet \cite{liang2017learning} & 2.25 & \textbf{3.40} & 2.44 & 2.07 & \textbf{2.76} & 2.19 & 0.12\\
CRL \cite{pang2017cascade} & 2.48 & 3.59 & 2.67 & 2.32 & 3.12 & 2.45 & 0.47\\
GC-Net \cite{kendall2017end} & 2.21 & 6.16 & 2.87 & 2.02 & 5.58 & 2.61 & 0.9\\
DRR \cite{gidaris2016detect} & 2.58 & 6.04 & 3.16 & 2.34 & 4.87 & 2.76 & 0.4 \\
SsSMNet \cite{zhong2017self} & 2.70 & 6.92 & 3.40 & 2.46 & 6.13 & 3.06 & 0.8\\
L-ResMatch \cite{shaked2016improved} & 2.72 & 6.95 & 3.42 & 2.35 & 5.74 & 2.91 & 48\\
Displets v2 \cite{guney2015displets} & 3.00 & 5.56 & 3.43 & 2.73 & 4.95 & 3.09 & 265\\
SGM-Net \cite{seki2017sgm} & 2.66 & 8.64 & 3.66 & 2.23 & 7.44 & 3.09 & 67\\
MC-CNN-acrt \cite{zbontar2016stereo} & 2.89 & 8.88 & 3.88 & 2.48 & 7.64 & 3.33 & 67\\
DispNetC \cite{mayer2016large} & 4.32 & 4.41 & 4.34 & 4.11 & 3.72 & 4.05 & 0.06\\
\hline
EdgeStereo (ours) & 2.27 & 4.18 & 2.59 & 2.12 & 3.85 & 2.40 &  0.27\\
\hlinew{0.75pt}
\end{tabularx}
\end{table}

\clearpage
%===========================================================
\bibliographystyle{splncs}
\bibliography{egbib}

%this would normally be the end of your paper, but you may also have an appendix
%within the given limit of number of pages
%\end{document}

%===========================================================

\end{document}